\documentclass[letterpaper, 10 pt, conference]{ieeeconf}  

\usepackage{graphicx}
\usepackage{amsmath}

\IEEEoverridecommandlockouts                              

\usepackage{color}
\usepackage{subfigure}

\usepackage{amssymb}
\usepackage{pifont}
\newcommand{\cmark}{\ding{51}}%
\usepackage{comment}

\usepackage{amsmath}
\usepackage{amssymb}
\usepackage{graphicx}
\usepackage{xcolor}
\usepackage{multirow}

\usepackage{bm}

\newcommand{\haolan}[1]{{}}

\title{\LARGE \bf
Safety-Critical Scenario Generation Via \\ 
Reinforcement Learning Based Editing
}

\author{Haolan Liu$^{1}$, Liangjun Zhang$^{2}$, Siva Kumar Sastry Hari$^{3}$ and Jishen Zhao$^{4}$
\thanks{$^{1}$Haolan Liu is a Ph.D. student with
        University of California San Diego, 9500 Gilman Dr, La Jolla, California
        {\tt\small hal022@ucsd.edu}}%
\thanks{$^{2}$Liangjun Zhang is with the Baidu Research,
        1195 Bordeaux Drive Sunnyvale, California
        {\tt\small liangjun.zhang@gmail.com}}%
\thanks{$^{3}$Siva Kumar Sastry Hari is with the NVIDIA Corporation,
      2788 San Tomas Expressway, Santa Clara, California
        {\tt\small shari@nvidia.com}}%
\thanks{$^{4}$Jishen Zhao is an associate professor at the
        University of California San Diego, 9500 Gilman Dr, La Jolla, California
        {\tt\small jzhao@ucsd.edu}}%
}

\begin{document}

\maketitle
\thispagestyle{empty}
\pagestyle{empty}

\begin{abstract}

Generating safety-critical scenarios is essential for testing and verifying the safety of autonomous vehicles. Traditional optimization techniques suffer from the curse of dimensionality and limit the search space to fixed parameter spaces. To address these challenges, we propose a deep reinforcement learning approach that generates scenarios by sequential editing, such as adding new agents or modifying the trajectories of the existing agents. Our framework employs a reward function consisting of both risk and plausibility objectives. The plausibility objective leverages generative models, such as a variational autoencoder, to learn the likelihood of the generated parameters from the training datasets; It penalizes the generation of unlikely scenarios. Our approach overcomes the dimensionality challenge and explores a wide range of safety-critical scenarios. 
Our evaluation demonstrates that the proposed method generates safety-critical scenarios of higher quality compared with previous approaches.
\end{abstract}

\section{Introduction}

The long-tail problem remains a significant hurdle in the development of 
autonomous vehicles (AVs)~\cite{Makansi_2021_ICCV, liu2019large}. To ensure safe navigation in real-world scenarios, autonomous vehicles need to 
handle a diverse range of rare and safety-critical corner cases. Due to the rarity of these scenarios in the real world, conducting effective and scalable safety testing and verification is challenging.
Furthermore, as modern autonomy stacks increasingly rely on data-driven approaches~\cite{DBLP:journals/corr/abs-2107-08142,bev-seg, liu2023interpretable}, they are becoming more vulnerable to corner cases that are poorly represented in training data.

Automatically generating safety-critical scenarios is crucial for addressing the long-tail problem. Previous studies 
treated the problem of generating adversarial scenarios as an optimization problem over the parameter space, using a specific cost function \cite{Rempe2021GeneratingUA, Wang_2021_CVPR,Wachi2019FailureScenarioMF}. The cost function is typically constructed based on factors such as closest distance or time-to-collision~\cite{10.5555/3327546.3327650,8793740}. The scenario parameters typically include the initial positions, orientations, trajectories of the traffic agents, lane topologies, and even different weathers. By considering a wide range of parameters, the generated scenarios can cover a diverse range of situations that 
are difficult to encounter in the real world, allowing for a more comprehensive AV safety validation.

\begin{figure}[]
    \center{
    \includegraphics[width=0.45\textwidth]{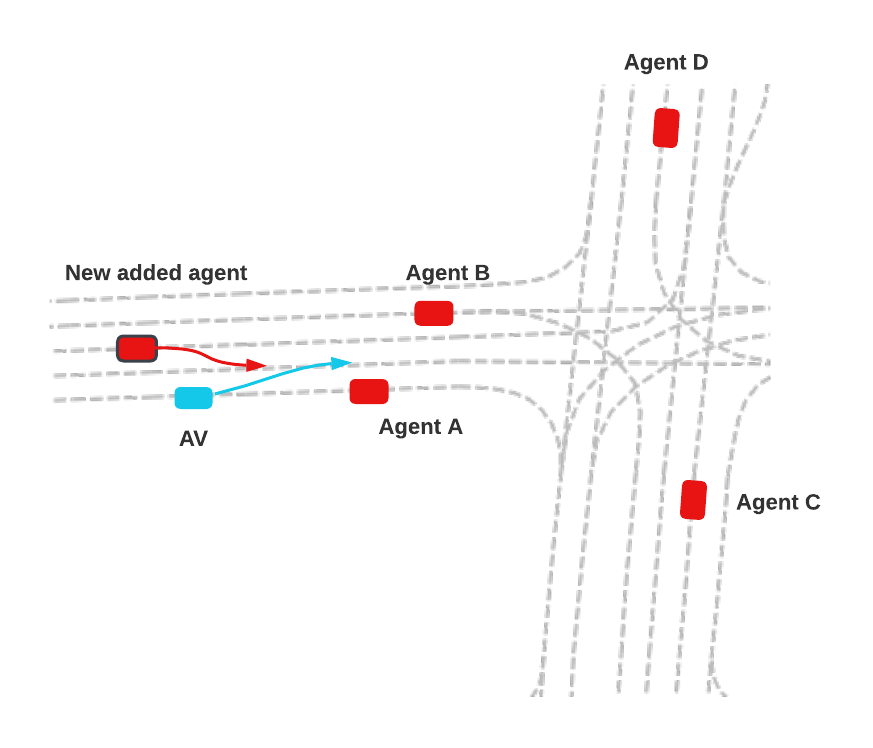}}
    \caption{(1) Adding exploration into scenario generation (newly added agents) helps generate high-quality safety-critical scenarios. (2) Jointly optimizing all the traffic agents (including agent $B,C,D$) in the scenario is not necessary, and can even hinder the optimization process, resulting in suboptimal results. Our framework edits the scenario efficiently in an agent-wise manner and search for desirable scenarios by sequential exploration. }
    \vspace{-20pt}

    \label{fig:motivation}
\end{figure}

As illustrated in Figure \ref{fig:motivation}, consider a scenario where the blue vehicle (AV) intends to change lanes to the left. In this case, the scenario parameter is the trajectory of traffic agent $A,B,C,D$. Previous studies have applied methods such as black-box optimization to iteratively search for perturbations that minimize the cost function, which is typically based on the closest distance metric. An optimal trajectory is one where the agent $A$ decelerates to reduce the closest distance, challenging the AV to safely change lanes.



However, such an approach only considers a fixed parameter space, which is the trajectory of the five agents, and does not take into account the diverse range of parameters that can affect the scenario. For instance, introducing a new agent into the scenario as in Figure~\ref{fig:motivation},  may not necessarily minimize the closest distance. However, such an "exploration" step will increase the scenario's complexity, making it easier to optimize in the long run. For example, agent $B$ can perform a lane change maneuver that intensifies the danger of the scenario. Moreover, this example also shows that previous manual cost function designs, e.g. closest distance are not general enough to accurately describe the risk of the scenario.



Furthermore, these approaches also face the challenge of high dimensionality. The widely-used black-box optimization techniques can only be applied to low-dimensional parameter spaces, typically with about 20 dimensions. To address this challenge, another branch of research has proposed optimizing scenarios in the latent space of a generative traffic model, using gradient-based methods~\cite{Rempe2021GeneratingUA}. However, such an approach requires performing joint optimization across all agents in the scenario, which may yield suboptimal results.




To address
the aforementioned challenges, we propose a scenario editing framework based on reinforcement learning, to generate safety-critical scenarios. Specifically, we employ a reinforcement learning agent to sequentially construct the scenario, by taking actions such as adding new agents or modifying the trajectory of existing agents. Our framework naturally supports exploration in scenario generation and automatically selects the most relevant traffic agents for generating safety-critical scenarios.

Our framework is trained to optimize a reward that consists of the risk objective and plausibility objectives.  
The risk objective quantifies the number of feasible driving plans for the AV, offering a detailed description of the risk in the scenario. 
The plausibility objectives apply a generative model to learn the likelihood of the generated parameters from the training datasets and penalize the generation of unlikely scenarios.
Besides, our approach can automatically identify the safety-critical agents.

Our 
approach leverages deep reinforcement learning (DRL) to overcome the curse of dimensionality that hampers traditional optimization techniques. 
Our method also allows us to generate 
scenarios with varying lengths of parameters, whereas previous approaches are restricted to fixed parameter spaces. 
The exploration-exploitation tradeoff in modern RL algorithms also facilitates the effective exploration of possible safety-critical scenarios.

%

We summarize our major contributions as follows:

\begin{itemize}
    \item We propose a reinforcement learning-based scenario editing framework, which can generate a diverse range of safety-critical scenarios by supporting operations such as adding agents, perturbations, and resampling. Our framework sequentially edits the scenario and efficiently explores the feasible scenarios. 
    \item We 
    design a generative model to learn the plausibility constraints of the scenario and propose a novel anchor-based risky score to describe desirable scenarios in 
    detail.
    \item 
    We perform a comprehensive analysis on our framework. Experiments demonstrate that the framework generates a diverse range of high-quality safety-critical scenarios.
\end{itemize}

\section{Related Work}

\paragraph{Adversarial Generation}

Adversarial scenario generation aims to identify scenarios with high risk by exploring various parameters based on a specified cost function. The cost function defines the ideal attributes of the scenarios, considering factors like the distance to nearby agents and the presence of undesirable sideway collisions, like sideway contact, as shown in \cite{8793740}. As this type of cost function can only be evaluated within non-differentiable simulators, previous research has used black-box optimization methods, such as Bayesian optimization (BO)~\cite{8793740} and evolutionary algorithms. However, these methods can become computationally demanding as the number of dimensions grows.


\paragraph{Generative Models}
Our work is also related to the generative model.
These models generate previously unseen scenarios by sampling from a model distribution that is similar to the data distribution. For example, DESIRE and HMSIM trains a conditional variational autoencoder to generate scenario-consistent trajectories for traffic agents~\cite{lee2017desire, liuhierarchical}. Similarly, TrafficGen trains an autoregressive model to sequentially insert new traffic agents~\cite{feng2022trafficgen}.

However, since safety-critical scenarios are rare in real-world datasets, the scenarios generated by the generative models are often normal and safe. To address this issue, previous work trained an ensemble of learned policies with generative adversarial imitation learning (GAIL) and leveraged cross-entropy methods to upweight the sampling frequency of rare risky events~\cite{10.5555/3327546.3327650}.
Another approach, called STRIVE, generates new scenarios by encoding existing scenarios into latent space and optimizing the scenario based on a differentiable planning model~\cite{Rempe2021GeneratingUA}.



\paragraph{Reinforcement Learning}

The reinforcement learning approach trains an adversarial policy to control scenario agents, with the goal of facilitating collisions with AVs. 
For instance, \cite{Wachi2019FailureScenarioMF} uses the Deep Deterministic Policy Gradient (DDPG) within a reinforcement learning framework to control surrounding agents and attack the AVs. 
The study designs an adversarial reward based on vehicle distance and adds a personal reward to discourage unrealistic agent behavior. The adversarial reward is allocated to each scenario agent based on their quantified contribution.

\paragraph{Structure Generation}
Our work also relates to structured data generation (e.g., graph or set structured data), such as the architecture of the neural network~\cite{pmlr-v70-jenatton17a, Lu2018StructuredVA}.
Such data exhibit conditional dependencies. For instance, the driving scenario can be defined by a large number of parameters, such as vehicle numbers, initial positions, and individual trajectories over time, which have complex interactions with each other. 

Previous works have used Bayesian Optimization (BO) methods on a predefined Euclidean space to generate safety-critical scenarios. However, black-box optimization methods are not practical for highly structured parameterization spaces, rendering the optimization process difficult.



\section{Proposed Method}

\begin{figure*}[t]
    \center{
    \includegraphics[width=0.9\textwidth]{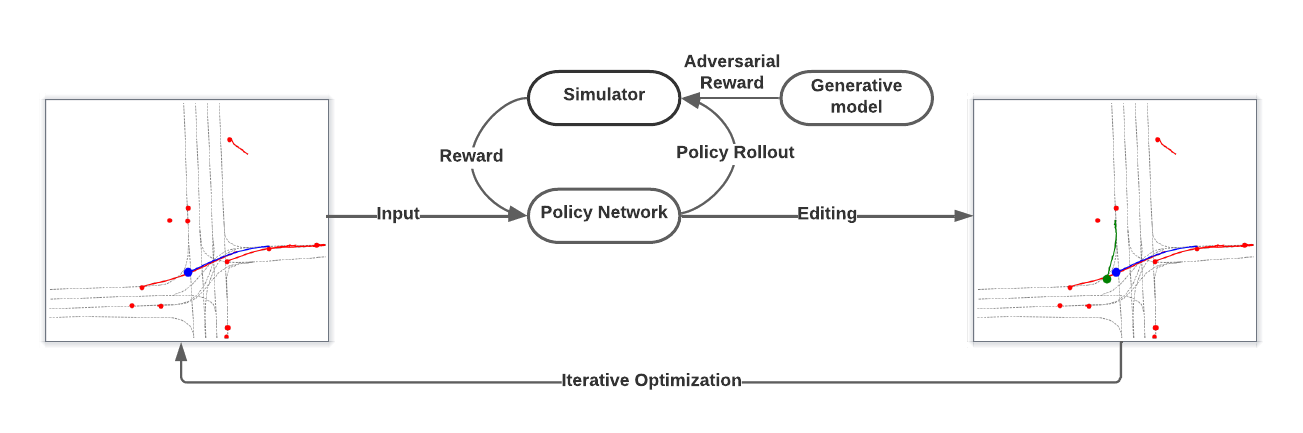}}
    \vspace{-0.1in}
    \caption{Our scenario editor is based on reinforcement learning, with a policy network that optimizes the scenario iteratively. In each iteration, the network takes the current scenario as input and outputs a distribution of editing actions. To train the agent, we use the risk model and adversarial loss from the pretrained model as the reward. In this example, the policy network adds a new agent to the scenario. The circle indicates the endpoint for each agent's trajectory. The blue, green, and red line indicate the trajectories of the ego vehicle, the newly added agent, and other agents in the scenario.
    }
    \vspace{-0.1in}
    \label{fig:overview}
\end{figure*}

\subsection{Problem Definition}
In this section, we formulate the scenario generation problem and discuss major challenges, as well as opportunities for improvement.
Given a scenario $x$, 
plausibility constraints $\mathbb{C}$, and a scoring function $s(x)$, our objective is to determine the optimal scenario, denoted by $x^*$:

\begin{equation}
    x^* = \operatorname*{argmin}_{x \in \mathbb{C}} s(x)
\end{equation}


The scoring function $s(x)$ represents the risk associated with driving scenario $x$; the scenario generation process must optimize this risk function while ensuring that the constraint $\mathbb{C}$ is not violated. An example of the scoring function $s(x)$ is the closest distance, as used by Abeysirigoonawardena et al.~\cite{8793740}. Minimizing the scoring function 
leads to closer agents in the scenario, increasing the possibilities of accidents.

Plausibility constraints $\mathbb{C}$ are typically categorized into two categories. (1) Soft constraints, which aim to ensure that the generated traffic agents are reasonably realistic. However, these constraints may be violated in safety-critical scenarios, where irregularities are present. (2) Hard constraints, which impose structural requirements on the generated traffic agents. For instance, the trajectory of a dynamic agent, such as a vehicle should be physically feasible.

To specify $\mathbb{C}$, previous works add penalty terms to the optimization objectives, such as traffic rule violation~\cite{8793740, Wang_2021_CVPR, Rempe2021GeneratingUA}. Another approach is limiting the allowable range of parameters. e.g., Wang et al. applies local search from existing scenarios by perturbing
the parameters, such as position and velocity profile, with limited magnitude~\cite{Wang_2021_CVPR}.
But such constraints also make the objective function $s(x)$ complex and hard to optimize.

To solve the constrained optimization problem, most prior approaches rely on black-box optimization strategies. At each timestep $i$, the strategy $M$ outputs $d_{i}$, which is iteratively improved to optimize the scoring function $s(x)$~\cite{Wang_2021_CVPR,8793740}. However, this approach works on a fixed and predefined parameter space and is subject to the curse of dimensionality. Moreover, there are no effective ways to explicitly control the plausibility of the output. To address these three challenges, we propose a scenario editing framework that utilizes deep reinforcement learning to edit the scenario, which is both flexible and scalable to high dimensions. Furthermore, we incorporate a pretrained generative model to explicitly define $\mathbb{C}$ during generation.


\subsection{Generation using RL} 


\noindent\textbf{Overview:} 
We represent the scenario as an unordered set ($V$, $T$), where the traffic agent matrix $ V \in {\mathbb{R}}^{n \times d_1} $  and the static context matrix $ T \in {\mathbb{R}}^{m \times d_2} $ are 
feature matrices assuming each element has $d_1$ or $d_2$ features, respectively.

Our framework formulates the scenario generation as an agent making sequential modifications to the scenarios, which is well-suited for reinforcement learning. The scenario editor iteratively edits the scenario representation by modifying the agent matrix $V$ or adding new features.
Figure~\ref{fig:overview} shows the process. At each time step, the policy network $\pi_\theta$=$p(a_t|s)$ takes the current scenario ($V$, $T$) and outputs an edit operation $a$. We then modify the scenario based on $a$ and compute the reward $R$ for the agent. 

We iteratively edit the scenario representation ($V$, $T$) as a Markov Decision Process (MDP) M = ($S$, $A$, $P$, $R$), where $S$ is a set of states of possible scenarios, $A$ is a set of actions that modify the scenario at each timestep. $P$ is the transition dynamics that describe how the scenario changes given an action.
$R$ is the specified reward function to describe the desired properties of the generated scenario. 


Our framework aims to optimize the expected cumulative reward $\sum_{t=0}^{\infty} \gamma^t r_{t}$, where $t$ denotes each timestep and $r_t$ denotes its reward. $\gamma$ is the discount factor that determines whether to prioritize the immediate return over the long-term return.
We set $\gamma=1$ in our framework. In addition, the maximum episode length $L_{max}$ bounds the number of edits to avoid overly complex scenarios.

\paragraph{\bf{State space}}

The state $S$ represents a fully observable scenario ($V$, $T$) where $T$ indicates the context, including details such as lanes, intersections, and traffic lights. 
$V$ indicates editable factors in the scenario, such as traffic agents' position and velocity.

{\paragraph{\bf{Action space}}

\begin{figure}[t]
    \center{
    \includegraphics[width=0.45\textwidth]{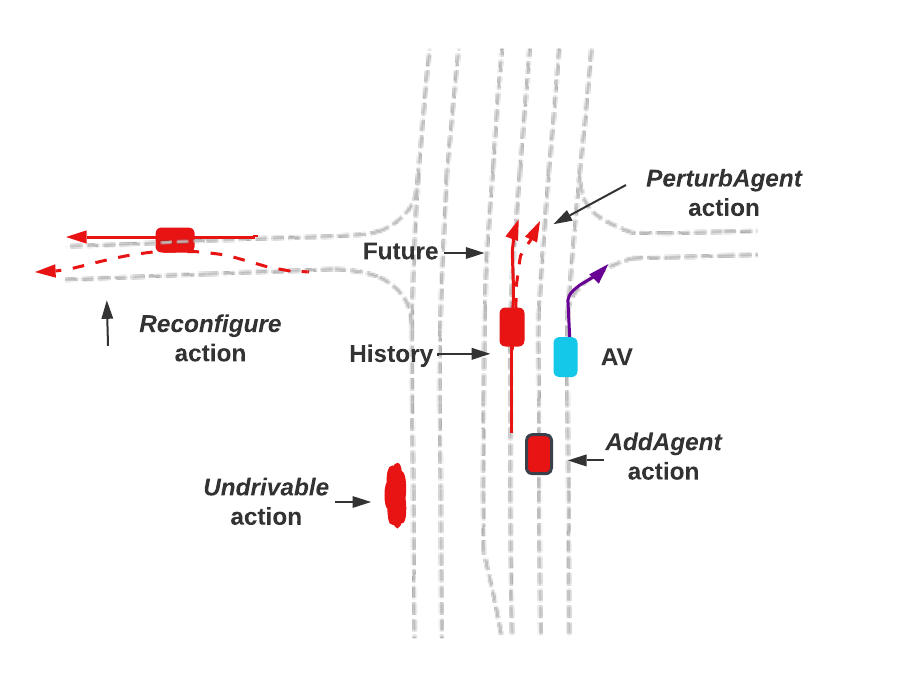}}
    \caption{The editing action supported by our framework. }
    \label{fig:action_space}
\end{figure}

The action describes how the scenario editor modifies the scenario. Figure~\ref{fig:action_space} shows the action that is supported by our framework. Specifically, we support the following actions:
\begin{itemize}
    \item \texttt{PerturbAgent}: perturb the trajectory of one of the existing agents based on its history and interaction with other agents. The continuous field includes the id of the selected agent and the perturbation vector to the original trajectory. 
    \item \texttt{AddAgent}: regressing the position, orientation, and trajectory of the newly added agents (we implemented vehicle only at this stage).
    \item \texttt{Reconfigure}: resampling a trajectory (including history and future trajectories) of the existing agent
    \item \texttt{Undrivable}: add an undrivable region to the map, which can represent parked vehicles or a construction area. The action also includes its coordinates. We use fixed-size regions.
    \item \texttt{Termination}: terminate the generation.
\end{itemize}

For generality, our action space is a discrete-continuous hybrid action space. 
\begin{equation}
    \hat{A} = { (f_{1:K},x_{1:K}) | f_k \in \mathcal{F}, x_k \in \chi_k, k \in [K] }
\end{equation}

Here $\mathcal{F} \in R$. $f_k$ indicates the discrete meta-action, it can be determined using argmax or sampling based on softmax operations. Once the discrete action $f$ is selected, e.g., adding a new agent, the rest of the action $x_{1:K}$ can be used to regress the initial position and trajectory
of the generated agents. 
In this way, our framework can support a flexible set of scenario editing operations.


Consider a case when the action space only includes the perturbation, our framework degrades to the previous adversarial generation approach~\cite{8793740, Rempe2021GeneratingUA, Wang_2021_CVPR}, which optimizes the representation ($V$, $T$) iteratively using black-box optimization. 
Therefore, our approach is a more generalized and expressive framework over 
prior work.

\paragraph{\bf{Transition Dynamics}}
With a given action $a$, the transition dynamics transform the scenario. We also perform necessary sanity checks for the generated action. For example, an action of adding a new agent, whose initial position collides with another vehicle, 
will be directly rejected.

\paragraph{\bf{Reward function}}
We apply two kinds of rewards to guide the search for safety-critical scenarios. (1) The final reward, which is only given after the termination actions, contains a risk model described in Section~\ref{subsec:risk}.
(2) Intermediate reward: the environment assigns an adversarial reward to penalize the unlikely behavior, by adding the data likelihood of each action given current scenarios. 
In addition, the environment also assigns a positive reward when the action passes the feasibility check, and assigns a negative reward otherwise.

\subsection{Policy Network} 

The policy network takes an intermediate scenario and outputs a stochastic action distribution.
At test time, we can sample or select the action with the maximum possibility.
We leverage the graph neural network to compute the embedding of the scenario, as in VectorNet~\cite{gao2020vectornet}. 
Specifically, the map and agent features (e.g., the lane geometry and the agent trajectory) are discretized into polylines, we use a graph neural network to extract polyline-level features and use a global self-attention layer to compute the embedding of the entire scenarios.
We first pretrain the graph neural network, on the motion forecasting task, and later optimize each scenario.
With the scenario embedding, we use a Multilayer Perceptron (MLP) network to compute the editing action, to generate the next scenario.



Our framework mainly employs Proximal Policy Optimization to train the policy network.
Proximal Policy Optimization (PPO) is a state-of-the-art on-policy algorithm, that leverages surrogate objectives to avoid undesirable big updates in the policy~\cite{DBLP:journals/corr/SchulmanWDRK17}.
We also pretrain the generative model for the adversarial reward, which will be further explained in \ref{sec:plausibility_control}.

\subsection{Risky Model}
\label{subsec:risk}
Previous studies have employed metrics, such as collision-or-not, time-to-collision, and closest distance, to quantify the risk associated with driving scenarios. However, these metrics are not always adequate in describing risk. For instance, in Figure~\ref{fig:motivation}, the addition of a new vehicle does not affect any of the aforementioned metrics, but it certainly makes the scenario more challenging to navigate correctly.


To overcome this limitation, we propose a risk model $r(x)$, which maps a scenario to a risk score based on the number of safe decisions available to the autonomous vehicle (AV). To achieve this, we discretize the feasible driving trajectory into $M$ predefined anchors, which can be obtained through clustering the trajectories in the dataset or using predefined grid targets. This approach is also used in anchor-based motion forecasting methods~\cite{Chai2019MultiPathMP}. Given a scenario $x$, we apply collision detection algorithms to the $M$ trajectories and compute the number of safe anchors. Figure~\ref{fig:risk_function} illustrates an example where $M=3$, and the predefined anchors include going straight, changing lanes to the right, and changing lanes to the left. We perform collision detection and determine that the two anchors shown in green\haolan{colored!} are safe.

For the generated scenario to be meaningful for testing the AV, it is essential that the AV has a way to avoid the collision. To prevent the generation of scenarios that are not "solvable"~\cite{9576023}, we have added a penalty term to the risk model in cases where no trajectory candidate is available.

\begin{figure}[t]
    \center{
    \includegraphics[width=0.4\textwidth]{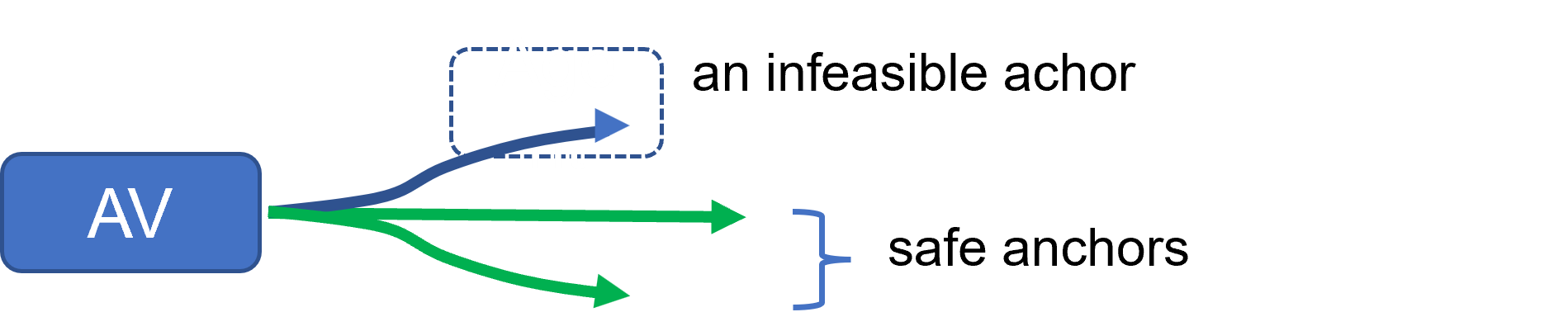}}
    \caption{The risk model evaluates the safety of the autonomous vehicle's driving plans. In the given scenario, there are three predefined plans: one blue and two green. The blue anchor leads to a collision, while the two green anchors are safe. The risk model counts the number of feasible driving plans and uses this information to guide the scenario editing process. }
    \label{fig:risk_function}
\end{figure}

\subsection{Plausibility Control}
\label{sec:plausibility_control}

We design an intermediate reward for each editing action, that penalizes unlikely behavior, such as generating a physically infeasible trajectory or placing a vehicle on the sidewalk. The previous approach designs heuristic rewards, such as penalty terms to the optimization objectives. Such heuristic design relies on human efforts and cannot scale to real-world complexities. 
We proposed an adversarial reward, that measures the deviation of the current scenario from the real-world scenarios.
Such rewards are used to guide the learning of the scenario editor.
In our paper, we implement the two kinds of adversarial reward, based on the data likelihood of the autoregressive model and the conditional variational autoencoder.
The conditional variational autoencoder (CVAE) is trained to maximize the conditional log-likelihood, by practically optimizing the variational lower bound:

\newcommand{\Lagr}{\mathcal{L}}

\begin{multline}
      \hat{\Lagr}_{C\!V\!\!A\!E}(\bf{x},\bf{y},\theta,\phi)=-\textit{KL}(q_{\phi}(\bf{z}|\bf{x},\bf{y})||p_{\theta}(\bf{z}|\bf{x}))+ \\ \sum_{L}^{i=1}log {p_\theta (\bf{y}|\bf{x},\bf{z^{(i)}})}  
\end{multline}

where L indicates the number of training samples, $z^{(i)}$ indicates each sample's latent code. $x$ and $y$ respectively indicates
the agent's future trajectory and its context information (map and other agents' history trajectories). The entire architecture includes a recognition network $q_\phi$ and a generation network $p_\theta$. Our recognition network also uses graph neural networks as the feature extractor. We pretrain the CVAE model using real-world datasets and use the reconstruction error as the plausibility score to penalize unlikely editing. For certain actions such as \texttt{AddAgent}, we add a rule-based common sense loss that forces the newly added agents to stay on lanes, and not too close to other agents, which are proved useful by previous works~\cite{Wang_2021_CVPR, Rempe2021GeneratingUA}.


We also apply a piecewise smooth function to normalize the data likelihood $\Lagr$, as it can become huge for some outliers. We split the reward value range into $K$ bins, and apply an individual $N_i$ ($i=1-K$) to normalize the reward. Such normalization can reduce the numerical instability for the training, and induce more smooth reward function.


\section{Experiments}

\subsection{Experimental Setup}

\paragraph{Dataset and Simulators}
We evaluate the proposed algorithm in a highway simulator (highway-env), which supports highly customizable scenario setup and also basic physical simulation~\cite{highwayenv}. The evaluated scenarios include multi-lane, intersection, and lane merging. 
We use the Argoverse Motion Forecasting Dataset~\cite{chang2019argoverse}, which contains over 300,000 tracked scenarios that span five seconds each, with a specific vehicle identified for trajectory forecasting. The dataset includes high-definition maps covering 290 km of mapped lanes, with both geometric and semantic metadata. We build a simple simulator that can import the Argoverse motion dataset, perform rollout of the AV planning policy, and also detect collision. The timestep frequency is 10 Hz.

\paragraph{Tested Systems}
For our tested AV system, we use two kinds of planners, (a) Reinforcement learning agent in the highway-env, trained by Proximal Policy Optimization (PPO)~\cite{DBLP:journals/corr/SchulmanWDRK17} (b) an imitation learning agent, based on ChauffeurNet~\cite{DBLP:journals/corr/abs-1812-03079} that exhibits good driving performance comparable to humans. 
We evaluate both planners across various scenarios and compute the average metric results.

\paragraph{Metrics}
To evaluate the quality of generated scenarios, we adopt a set of several 
metrics, which is a common practice used in previous works~\cite{Wang_2021_CVPR}. 1) \textbf{Collision rate}, which represents the percentage of scenarios in which the autonomous vehicle (AV) collides with another actor. This metric allows us to determine if the generated scenario can effectively identify potential safety issues with the AV planner. aggressive way. 2) \textbf{L2 distance} to the imitation learning planner's trajectory as a measure of how closely the generated trajectories align with those of human experts. This metric helps us evaluate the deviation of the generated trajectories from human driving behavior.
3) The \textbf{Distribution JSD} metric. We compute the distribution of velocity, acceleration, and jerk (derivative of the acceleration)
in the datasets, and discretize them into histograms. We compute the Jensen-Shannon Divergence between the generated scenarios and dataset scenarios~\cite{Suo_2023_CVPR}.
This metric indicates the motion similarities between real-world scenarios and generated scenarios.
4) we also introduce the \textbf{Traffic Rules Violation} metric, quantifying the frequency of generating behaviors such as off-road driving.

\subsection{Generation Quality Evaluation}
We evaluate the generated scenario by running our tested AV planner in the scenario.
To establish a baseline, we compared our work with Bayesian Optimization~\cite{8793740}, AdvSim~\cite{Wang_2021_CVPR}, and STRIVE~\cite{Rempe2021GeneratingUA}. STRIVE applies a preprocessing step to filter out scenarios that are unlikely to result in a collision~\cite{Rempe2021GeneratingUA}. To ensure a fair comparison, we did not apply this preprocessing step for any of the three algorithms.
Using the same scenario from the dataset, we evaluated the performance of the four algorithms by optimizing until convergence.
We also filter out some physically infeasible scenarios using the bicycle model and also rule-checking.

\begin{table*}[]
\centering
\caption{Evaluation of the collision rate and the plausibility metrics across previous works and our work.}
\label{tab:eval}
\begin{tabular}{ccccccc}
\hline
         & Risk & \multicolumn{5}{c}{Plausibility} \\ \hline
 & Safety                 & Dataset Reconstruction & \multicolumn{3}{c}{Distribuition JSD} & Common Sense                    \\ \hline
 & Collision Rate(Percent)$\uparrow$  & L2 Distance$\downarrow$            & Velocity$\downarrow$      & Accel$\downarrow$       & Jerk$\downarrow$      & Traffic Rules Violation(Percent)$\downarrow$  \\ \hline
BO\cite{8793740}  & 1.1  & 9.81  & 0.83 & 0.61 & 0.62 & 8.4 \\ \hline
AdvSim\cite{Wang_2021_CVPR}   & 3.7  & 3.44  & 0.65 & 0.49 & 0.66 & 5.7 \\ \hline
STRIVE\cite{Rempe2021GeneratingUA}   & 4.3  & 1.93  & 0.21 & 0.17 & 0.35 & 2.1 \\ \hline
Our work & \textbf{7.8}  & \textbf{1.72}  & \textbf{0.17} & \textbf{0.18} & \textbf{0.23} & \textbf{1.6} \\ \hline
\end{tabular}
\end{table*}

\begin{figure}
    \centering
    \subfigure{\includegraphics[width=0.2\textwidth]{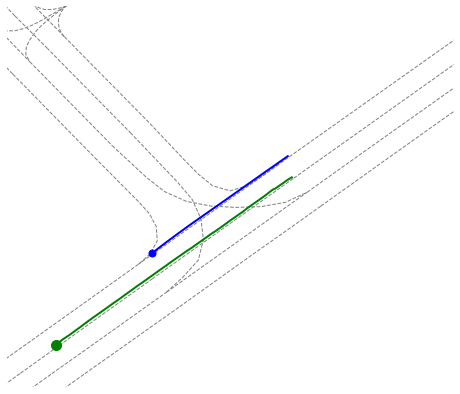}} 
    \subfigure{\includegraphics[width=0.2\textwidth]{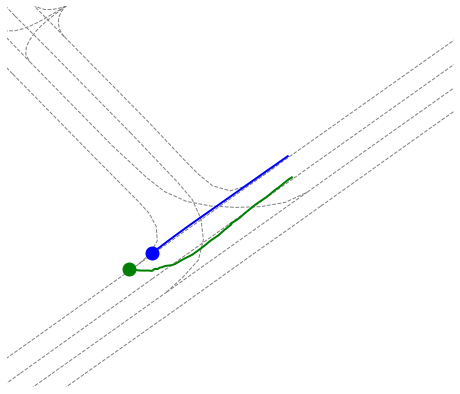}} 
      \subfigure{\includegraphics[width=0.22\textwidth]{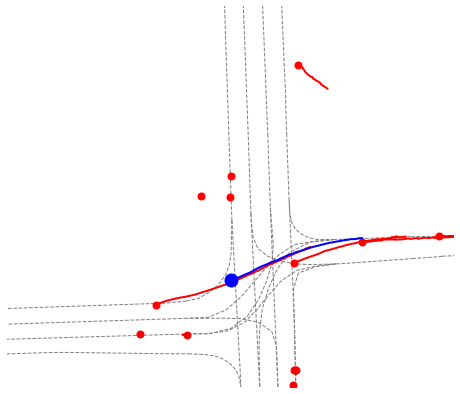}}
    \subfigure{\includegraphics[width=0.22\textwidth]{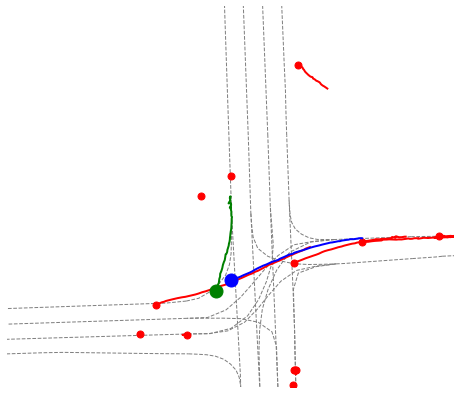}}
    \caption{ The original scenario (left) and our edited version (right). The dotted line represents the lane centerline, and the circle denotes the final point of the trajectory. The green and red lines represent the trajectories of traffic agents, with the green line highlighting the safety-critical agents. For clarity, we only show the nearby agents in the figure. The blue line represents the AV's trajectory. (1) lane changing. (2) intersection. }
    \label{fig:vis}
\end{figure}

\paragraph{Risk Evaluation} We show the collision metric in Table~\ref{tab:eval}. 
We optimize the existing scenario from the dataset. 
Our experiments demonstrate that our approach effectively generates challenging scenarios for various systems, including our imitation learning and reinforcement learning planners. This results in an average collision rate of 7.8\%, which is over 80\% higher compared to the original state-of-the-art method, STRIVE.

\paragraph{Plausibility Evaluation}
Table~\ref{tab:eval} also shows the plausibility metrics evaluated on the generated scenarios. We can see that the Bayesian Optimization (\textbf{BO} and \textbf{AdvSim}) approach exhibits unrealistic scenarios, compared with STRIVE and our approach, which can be mainly explained by the strong data prior in both models. Our approach achieves better plausibility in every metric compared with previous works, while also leading to more challenging scenarios (higher collision rates).

\begin{figure}[!htb]
    \centering
    \begin{minipage}{0.23\textwidth}
        \centering
        \includegraphics[width=\textwidth]{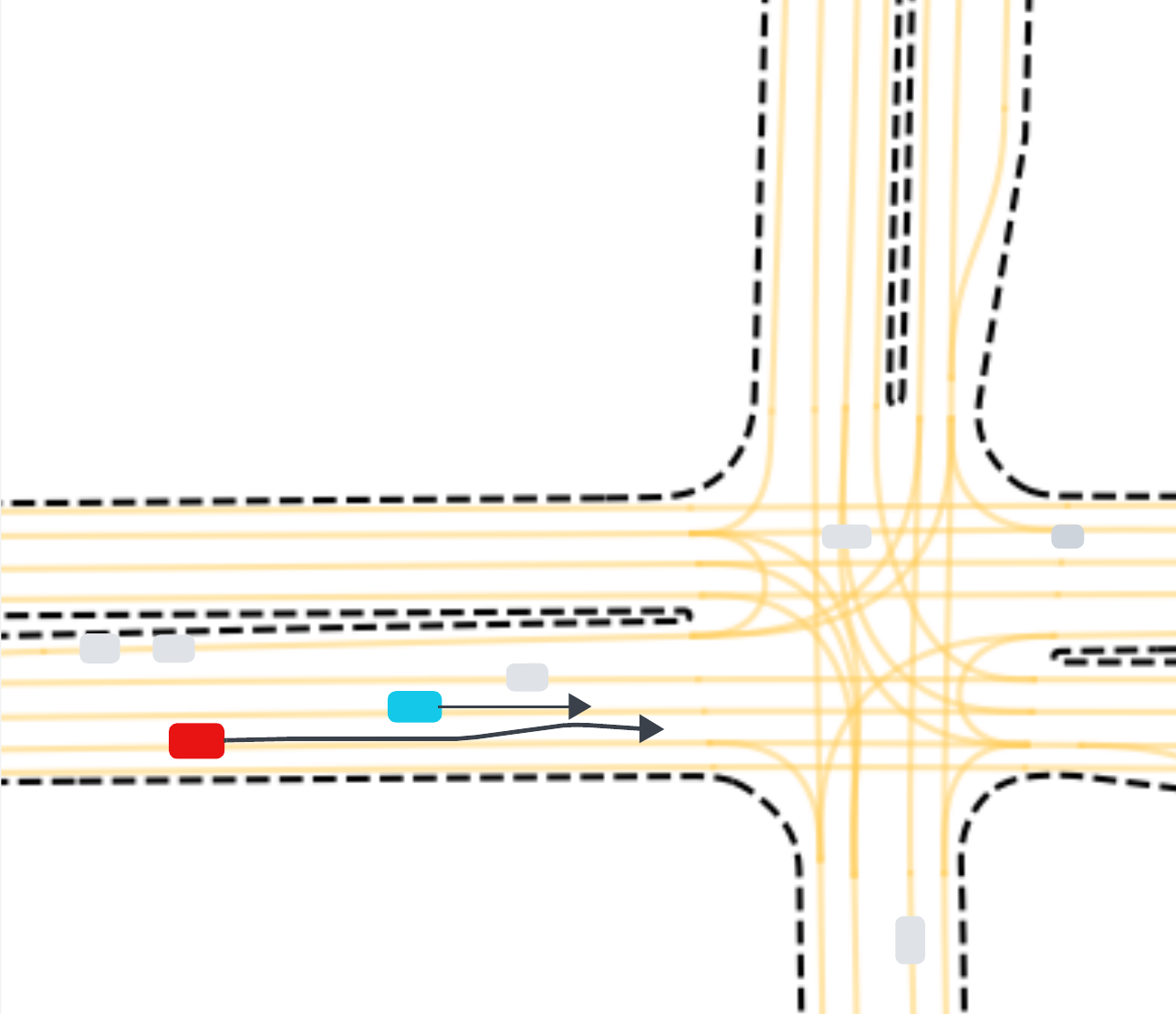}
      
    \end{minipage}
    \hfill
    \begin{minipage}{0.23\textwidth}
        \centering
        \includegraphics[width=\textwidth]{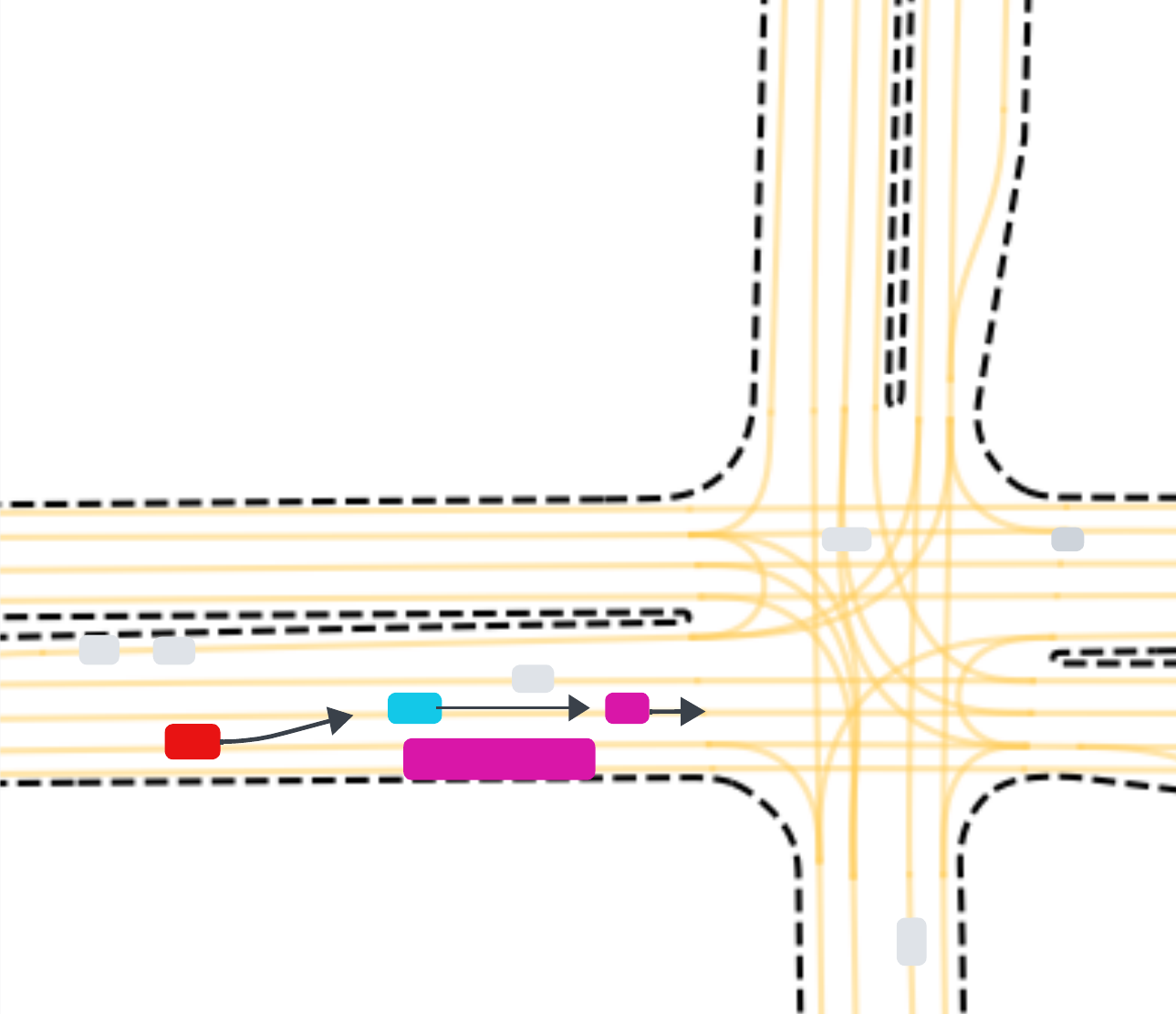}
        
    \end{minipage}
    \caption{The blue vehicle indicates the ego vehicle (AV), the red vehicle indicates the original attacker vehicle and the grey vehicles indicate other agents in the scenario. STRIVE generates plain cutting-in behaviors, while our framework adds another agent (purple vehicle) and also an undrivable region (a purple parked bus).}
        \label{fig:comparison}
\end{figure}
\subsection{Scenario Visualization}

In Figure~\ref{fig:vis}, we compare the original scenario with the edited version to demonstrate the diversity of safety-critical scenarios generated by our editing policy, which includes lane-changing and intersection scenarios. In the first example, our policy adds a new agent to the scenario and performs a lane change in the intersection. In the second example, the agent of interest's trajectory in an intersection is resampled and perturbed to a safety-critical lane merging scenario. 

We also compare the scenario generated by our framework and STRIVE in Figure~\ref{fig:comparison}.
Prior methods such as STRIVE can only generate risky behaviors from existing agents, leading to the cutting-in behaviors in the left figure. Our framework can add new agents or obstacles (the purple vehicle and the parked bus) and also reconfigure existing agents (red vehicle), to generate more natural but still challenging scenarios.
These examples show the effectiveness of our framework in generating a wide range of realistic and complex safety-critical scenarios.

\subsection{Ablation Study}
We conduct an ablation study using various configurations (M0-3) to validate the effectiveness of our design choices. The results indicate that the \texttt{AddAgent} and \texttt{Reconfigure} actions not only increase the risk level of generated scenarios but also enhance their plausibility. Additionally, our risk model proves effective in guiding the search for riskier scenarios.

\begin{table}[]
\caption{Ablation study of action space design and the risk model. M0 indicates a vanilla RL baseline.}
\label{tab:ablation}
\begin{tabular}{cccccc}
\hline
 &
  \begin{tabular}[c]{@{}c@{}}AddAgent\\ Action\end{tabular} &
  \begin{tabular}[c]{@{}c@{}}Reconfigure\\ Action\end{tabular} &
  \begin{tabular}[c]{@{}c@{}}Risk \\ Model\end{tabular} &
  \begin{tabular}[c]{@{}c@{}}Collision \\ Rate/Percent\end{tabular} &
  \begin{tabular}[c]{@{}c@{}}Plausibility\\ Score\end{tabular} \\ \hline
M0 &       &       &       & 3.2 & 7.14 \\ \hline
M1       & \cmark &       &       & 5.6 & 6.83 \\ \hline
M2       & \cmark & \cmark &       & 7.1 & 7.52 \\ \hline
M3       & \cmark & \cmark & \cmark & 7.8 & 7.44 \\ \hline
\end{tabular}
\end{table}

\section{CONCLUSIONS}
In this work, we proposed a scenario editing framework that leverages reinforcement learning to perform a set of operations on the driving scenario, including adding agents, perturbation, and resampling. We use generative models to learn the plausibility constraints of the scenarios and a novel anchor-based risky score to describe the desirable scenarios in a more fine-grained manner. Our evaluation shows that our framework can generate high-quality safety-critical scenarios, outperforming previous methods.

In future work, we plan to explore how our framework can be extended to handle more complex scenarios, such as those involving pedestrians or cyclists. We also plan to investigate how our framework can be integrated with existing scenario generation methods for autonomous vehicles.

\addtolength{\textheight}{-12cm}   









\clearpage

\bibliographystyle{abbrv}
\bibliography{references}

\end{document}